\documentclass[journal]{IEEEtran} % removed 'compsoc' for arXiv

\usepackage{amsmath}
\usepackage{graphicx} % Handles PDF/LaTeX graphics
\usepackage{cite} % IEEE-style citations
\usepackage{amssymb}
\usepackage{array}
\usepackage{url}
\usepackage{booktabs} % For tables
\usepackage{makecell} % Multi-line table cells
\usepackage{float}    % [H] placement specifier
\usepackage{caption}  % Advanced captions
\usepackage{xcolor}   % Colored text

\begin{document}

% --- CLEAR HEADER ---
\markboth{}{} % removes default IEEE headers

% --- TITLE, AUTHORS, AND ABSTRACT ---
\title{WaveNet's Precision in iEEG Classification}

\author{Casper~David~van~Laar and Khubaib~Ahmed
\thanks{K. Ahmed is with the University of Wolverhampton, U.K.}}

\IEEEtitleabstractindextext{%
\begin{abstract}
This study introduces a WaveNet-based deep learning model designed to automate the classification of Intracranial Electroencephalography (iEEG) signals into physiological activity, pathological/epileptic activity, power-line noise, and other non-cerebral artifacts categories. Traditional methods for iEEG signal classification, which rely on expert visual review, are becoming increasingly impractical due to the growing complexity and volume of iEEG recordings. Leveraging a publicly available annotated dataset from Mayo Clinic and St. Anne's University Hospital, the WaveNet model was trained, validated, and tested on 209,231 samples with a 70/20/10 \% split. The model achieved a classification accuracy exceeding previous non specialized CNN and LSTM-based approaches, and was benchmarked against a Temporal Convolutional Network (TCN) baseline. Notably, the model achieves high discrimination of noise and artifact classes (precision = 0.98 and $\approx 1$, respectively). Although, classification between physiological and pathological signals exhibits a modest but clinically interpretable overlap, with F1-scores of 0.96 and 0.90 and 175 and 272 cross-class false positives, respectively. Reflecting inherent clinical overlap. WaveNet's architecture, originally developed for raw audio synthesis, is well-suited for iEEG data due to its use of dilated causal convolutions and residual connections, enabling it to capture both fine-grained and long-range temporal dependencies. The research also details the preprocessing pipeline, including dynamic dataset partitioning, the use of focal loss to combat class imbalances and normalization steps that support model high performance. While results demonstrate strong in-distribution performance, generalizability across datasets and clinical settings has yet to be established.
\end{abstract}

}

\maketitle
\IEEEdisplaynontitleabstractindextext

% --- START OF MAIN CONTENT ---
% Your sections go here

% --- SECTIONS ---

\section{Introduction}
The classification of Intracranial Electroencephalography (iEEG) data into physiological, pathological, artifact, and noise categories is essential for accurate neurophysiological analysis and clinical diagnostics. Traditional methods rely heavily on expert visual review, which is not scalable given the increasing volume and complexity of iEEG recordings. Trained iEEG technologists are crucial in identifying abnormalities and enhancing patient outcomes. Manual EEG review by neurophysiologists and technologist is prone to errors, non-scalability, subjectivity, and inconsistencies due to varying expertise, fatigue, and cognitive biases \cite{leso2021shift}. In contrast, automated models \cite{nejedly2019artifact, nejedly2019graphoelements} are scalable, mitigate oversight, and use deep learning (DL) to discern subtle iEEG patterns, improving reliability for precise diagnoses and treatments.

\begin{figure}[H]
    \centering
    \includegraphics[width=0.8\columnwidth]{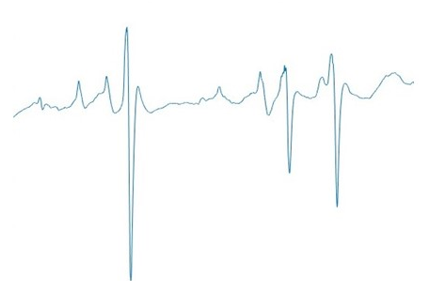}
    \caption{Healthy physiological signals. Characterized by a steep incline and decline before and after the peak.}
    \label{fig:eeg_physiological}
\end{figure}

\begin{figure}[H]
    \centering
    \includegraphics[width=0.8\columnwidth]{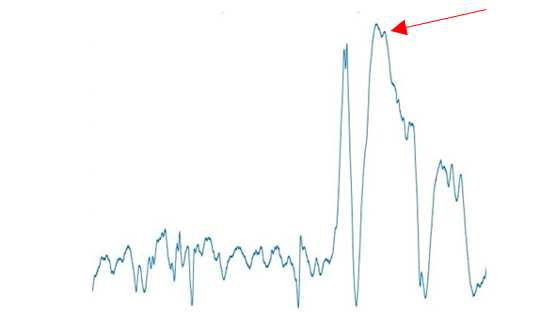}
    \caption{Example of abnormal brain activity observed in epilepsy-prone neurons identified by a sharp wave transient. High frequency oscillation (HFO) observed atop the spike peak.}
    \label{fig:eeg_pathological}
\end{figure}

A Temporal Convolutional Network (TCN) is a type of DL neural network designed for processing sequential data. WaveNet is a specific type of TCN that excels at handling the intricate temporal characteristics of audio waveforms due to its architecture, which includes dilated causal convolutions and residual connections \cite{oord2016wavenet}. In WaveNet, causal convolutions ensure that the output at a time step depends on the current and previous time steps, mimicking the autoregressive nature of time series data like iEEG signals. This is crucial as it prevents information leakage from future time steps into the current prediction.

iEEG data are continuous signals reflecting the brain's electrical activity, exhibiting complex temporal dependencies and dynamic variations \cite{islam2023editorial}. The architecture is particularly well-suited for this task, as it can capture both fine-grained and long-range temporal dependencies in the data. This capability makes WaveNet a strong option for iEEG data classification:
\begin{itemize}
    \item \textbf{Temporal Dynamics:} WaveNet’s ability to model long-range dependencies and temporal patterns aligns well with capturing iEEG signals' temporal dynamics. iEEG signals fit this description perfectly, as they exhibit complex temporal dependencies and variations (see Figure~\ref{fig:eeg_physiological} and \ref{fig:eeg_pathological} for examples).
    \item \textbf{Hierarchical Features:} WaveNet’s hierarchical nature allows it to learn multi-scale features, crucial for identifying patterns in iEEG data at different temporal resolutions. High-resolution features are captured by lower layers, while higher layers capture long-term, low-resolution patterns. This hierarchical feature extraction is crucial for modeling complex temporal dynamics in sequential data, which can allow the model to differentiate between physiological and pathological iEEG data (Figures~\ref{fig:eeg_physiological} and \ref{fig:eeg_pathological}) and noise or artifacts.
    \item \textbf{High Resolution:} WaveNet can capture detailed variations in iEEG signals, essential for accurate classification, like the differences in the peaking behaviour of the iEEG signals of pathological and healthy patients (Fig.~\ref{fig:eeg_physiological} and Fig.~\ref{fig:eeg_pathological}).
\end{itemize}
Therefore, we propose using a WaveNet model to automate the classification of iEEG data. To provide context, we also benchmark its performance against a TCN baseline. WaveNet’s architecture makes it well-suited for capturing the complex temporal patterns in iEEG signals. Automating iEEG data classification with WaveNet will not only enhance efficiency and accuracy but also ensure more consistent and reliable diagnostics, ultimately improving patient outcomes and advancing the field of neurophysiology.
\begin{itemize}
    \item \textbf{Temporal Dynamics:} Long-range dependencies and temporal patterns in iEEG.
    \item \textbf{Hierarchical Features:} Multi-scale features, distinguishing physiological, pathological, and artifact signals.
    \item \textbf{High Resolution:} Detailed waveform variations crucial for precise classification.
\end{itemize}

\section{Related works}
Accurate detection of epileptiform spikes is essential for the diagnosis and clinical management of epilepsy. Earlier methods relied heavily on manual annotation or rule-based algorithms, which are time consuming and subject to inter-rater variability, while recent deep learning approaches show promise in automating (i)EEG spike classification by learning features directly from raw or minimally processed signals \cite{bai2025epileptic}.

Initial work applied convolutional neural networks (CNNs) to this task. Johansen et al. \cite{johansen2016epileptiform} reported a CNN-based classifier with an area under the curve (AUC) of 95 \%, effectively identifying epileptiform spikes with high sensitivity and low false positive rates. Nejedly et al. \cite{nejedly2019graphoelements} extended this approach by combining graphoelement-derived features with both CNN and long short-term memory (LSTM) architectures, achieving F1 scores of 0.81 and 0.80, respectively. Their specialized patient tuned CNN model for pathological activity classification reached an average F1 score of 0.96. They also explored artifact rejection in iEEG using CNN-based classifiers \cite{nejedly2019artifact}, improving reliability by filtering out physiological noise from the input signals.

Subsequent research incorporated attention mechanisms into recurrent models to better capture temporal dependencies. Fukumori, Yoshida, and Tanaka \cite{fukumori2024spike} proposed a self-attentive recurrent neural network that dynamically emphasized relevant time steps, achieving an average accuracy of 90 \% in epileptic spike detection. Similarly, Rukhsar and Tiwari \cite{rukhsar2024arnn} introduced an Attention Recurrent Neural Network (ARNN) that applied self-attention across multi-channel iEEG data. Evaluated on datasets like CHB-MIT and canine iEEG from the Mayo Clinic and North Carolina State University, the model achieved an average F1 score of 0.94 and outperformed LSTM and Vision Transformer baselines. However, ARNN remains a specialized model trained and evaluated within specific datasets, and its performance on cross-patient or cross-institution data remains an open question.

These developments reflect a broader trend toward hybrid deep learning architectures that combine spatial and temporal modeling. The integration of CNNs, RNNs, and attention mechanisms has improved classification accuracy and continues to shape efforts toward real-time and/or scalable EEG analysis.

\section{Dataset}
The dataset for this article comprises annotated  iEEG recordings from Mayo Clinic (MN, USA) and St. Anne’s University Hospital (Brno, Czech Republic). The dataset includes iEEG data labelled into four categories: physiological activity, pathological/epileptic activity, power-line noise, and other non-cerebral artifacts. Each data sample of 3 seconds is stored in a .mat file with a sampling frequency of 5kHz and consists of a vector with dimensions $(1,15000)$ containing a total of 209,231 samples. Comprehensive instructions for dataset usage are provided in the associated scientific papers \cite{nejedly2019artifact}\cite{nejedly2019graphoelements}.

\section{Methods}
\subsection{Preprocessing steps}
The preprocessing steps involved loading iEEG data from .mat files stored in directories, and organizing them into HDF5 format. This format optimizes storage and access speed while handling large datasets efficiently. After this preprocessing step, the data is dynamically split into distinct training, validation, and test sets. This splitting is achieved through random sampling individual samples, ensuring separation and preventing data leakage between the sets during model evaluation and training. The entire pipeline is designed with mechanisms to maintain the integrity and separation of data throughout each stage of processing. Then, the input layer normalizes the input using Z-score normalization, adding $\varepsilon = 1 \times 10^{-8}$ to prevent division by zero.

\subsection{Model structure}
The WaveNet model is built using dilated causal convolutions to capture long-range dependencies in iEEG signals. It employs seven sequential residual blocks with increasing dilation rates of $[1, 2, 4, 8, 16, 32, 64]$, allowing the receptive field to grow exponentially while maintaining the temporal ordering for time series modeling. 

Each residual block includes:
\begin{itemize}
    \item \textbf{Swish activation function} defined as $\text{Swish}(x) = x \cdot \text{sigmoid}(x)$.
    \item A dilated convolution layer with 32 filters and a kernel size of 3. For each convolutional layer with dilation rate $r$ and kernel $k$ size 3:
    \[
    z_i(l) = \text{swish}\left(\sum_{j=0}^{k-1} w_j^{(l)} * x_{(i-j)}^{(l-1)}\right)
    \]
    \item A residual connection,
    \item and A skip connection that feeds into a cumulative output stream.
\end{itemize}

These skip connections are summed and passed through two additional $\text{Conv1D}$ layers. The first with 32 filters, and the second with 4 filters (equal to the number of output classes). All before being globally averaged across the time dimension via a $\text{GlobalAveragePooling1D}$ layer. A final softmax activation layer produces the class probability distribution for each iEEG segment.

To combat overfitting, an novel adaptive dropout mechanism is employed. This starts at an initial rate of 0.20 and dynamically adjusts throughout training based on a composite score combining validation accuracy, AUC, loss, and the relative gap between training and validation performance (for more detail see appendix A). The final observed dropout rate converged to 0.23 by the end of training. 

Optimization is handled by the Adam \cite{kingma2015adam} optimizer with an initial learning rate of 0.001, and L2 regularization with $\lambda=0.0001$ is applied to constrain model complexity and enhance generalization. To address class imbalance, particularly the dominance of physiological and pathological iEEG patterns, the training pipeline applies dynamic class weighting called focal loss\cite{lin2018focal}. When using focal loss, class weights inversely proportional to class frequencies are passed as alpha parameters, focusing learning on underrepresented classes and down-weighting well-classified samples. 

WaveNet models the conditional probability distribution over sequential data, estimating the joint probability of a sequence $P(x) = \prod_{t=1}^{T} P(x_t | x_1, \dots, x_{t-1})$. This autoregressive structure makes might make WaveNet well-suited for modeling EEG signals, which exhibit strong temporal dependencies and nonstationary dynamics.

\subsection{Baseline: Temporal Convolutional Network (TCN)}
The Temporal Convolutional Network (TCN) was introduced as an additional baseline to provide a convolutional, non-autoregressive counterpart to WaveNet. While both architectures employ causal, dilated convolutions and residual connections, the TCN is tailored for classification rather than autoregressive synthesis. Its design kept receptive field and capacity comparable to WaveNet.
\textbf{Architecture:}
\begin{itemize}
  \item Residual stack of 8 causal Conv1D blocks with dilation rates $[1,2,4,8,16,32,64]$, kernel size $2$, and 8 filters per layer.
  \item Each block contains two Conv1D layers with ReLU activations and weight normalization; residual paths use $1 \times 1$ projections when needed.
  \item Fixed dropout of $0.005$ within each block and Layer Normalization applied after residual addition.
  \item Classification head: two Conv1D layers (8 then 4 filters), GlobalAveragePooling1D, and softmax output.
\end{itemize}

\textbf{Training regimen:}
\begin{itemize}
  \item Dataset: identical HDF5 source as WaveNet with 70/20/10 (train/validation/test) split.
  \item Optimizer: Adam ($\text{lr}=1 \times 10^{-3}$) with L2 regularization ($\lambda = 1 \times 10^{-4}$).
  \item Loss: focal loss ($\gamma = 2.0$) with class-weighted $\alpha$ parameters to correct imbalance.
  \item Batch size: 16.
  \item Early stopping on validation AUC (patience = 3) and model checkpointing based on macro-F1.
\end{itemize}
The baseline was implemented and trained using identical preprocessing and evaluation pipeline as WaveNet to ensure comparability. The core aim was to evaluate whether a purely convolutional, dilation-based classifier could rival WaveNet’s balance of temporal context modeling and residual learning.

\subsection{Training summary}
The WaveNet and TCN models were trained on different hardware platforms but shared a common optimization strategy. Both models employed a custom focal loss function \cite{lin2018focal} to address class imbalance and used Adam optimization with dropout to mitigate overfitting.

WaveNet was trained on the Kaggle platform using a Tesla P100 GPU with optimized hardware acceleration and adaptive dropout, with batch size, learning rate, 36,710 trainable parameters and 2 non trainable parameters (see Appendix A for full details).

TCN training was conducted locally on an MSI Katana laptop equipped with an NVIDIA GeForce RTX 3050 Laptop GPU using mixed-precision acceleration; fixed dropout (0.005) was applied within residual blocks, and dilation rates, batch size (16), learning rate (1e-3), and number of filters (8) were empirically optimized. It contained 4,852 parameters in total, of which 4,402 are trainable.

\section{Results}
The WaveNet model was trained on a curated dataset from St. Anne’s University Hospital and the Mayo clinic consisting of 209,231 EEG samples, split into 70\% training, 20\% validation, and 10\% testing subsets. The model achieved an average F1 score of 0.94, outperforming prior methods such as the general CNN approach \cite{nejedly2019artifact} and the graphoelements Convolutional LSTM model \cite{nejedly2019graphoelements} which obtained an average F1 scores of 0.80 and 0.81 respectively (see Table~\ref{tab:classification_performance}). While underperforming on the specialized model which reached a F1 of 0.96.

\begin{figure}[!t]
    \centering
    \includegraphics[width=0.8\columnwidth]{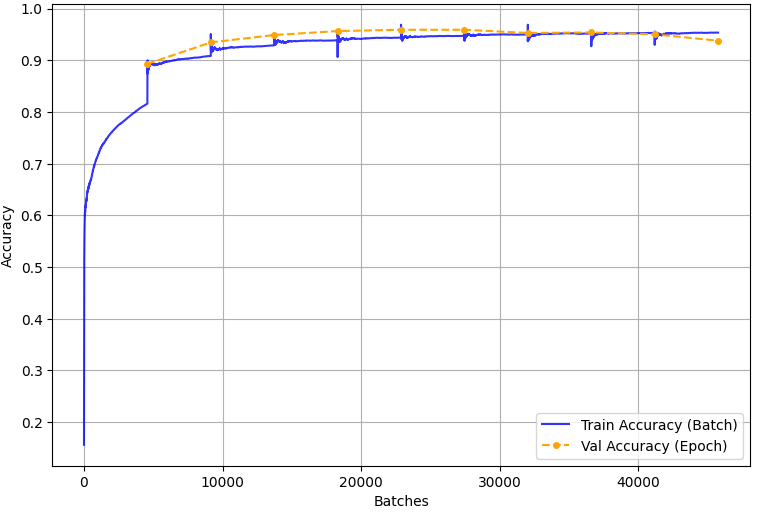}
    \caption{Wavenet accuracy figure, plotted every 150 batches each batch contains 32 samples for the training set. Yellow for the validation set plotted every epoch Minor spikes at epoch boundaries reflect the “epochal sawtooth phenomenon” \cite{liu2025epochal}, caused by data reshuffling and optimizer dynamics.}
    \label{fig:ACC_plot}
\end{figure}

\begin{figure}[!t]
    \centering
    \includegraphics[width=0.8\columnwidth]{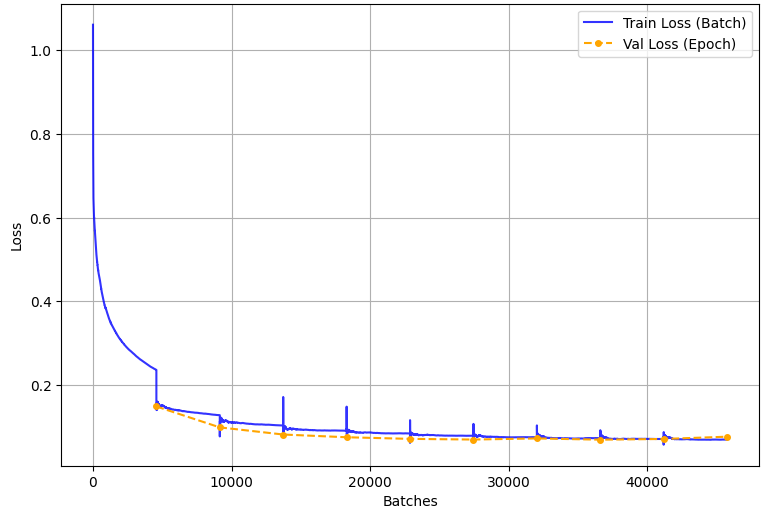}
    \caption{Wavenet validation and training loss figure, plotted every 150 batches each batch contains 32 samples for the training set. Yellow for the validation set plotted every epoch. Transient spikes at the start of epochs correspond to the same Epochal Sawtooth phenomenon as explained by liu et al(2025).}
    \label{fig:LOSS_plot}
\end{figure}

Training dynamics (Figure~\ref{fig:ACC_plot}) demonstrate that the WaveNet reached ~95\% accuracy on both the training and validation set by the third epoch, stabilizing around 95\%. The validation loss and accuracy curves (Figure~\ref{fig:LOSS_plot}) confirm stable convergence without significant overfitting.

\begin{table}[H]
    \centering
    \caption{Confusion Matrix for WaveNet Model Predictions on Test Set}
    \label{tab:WAVE_matrix}
    \small
    \begin{tabular}{lcccc}
        \toprule
        \textbf{True/Pred} & \textbf{Noise} & \textbf{Artifacts} & \textbf{Physio.} & \textbf{Patho.} \\
        \midrule
        \textbf{Noise} & 3561 & 1 & 0 & 1 \\
        \textbf{Artifacts} & 10 & 3372 & 371 & 47 \\
        \textbf{Physiological} & 0 & 45 & 10866 & 175 \\
        \textbf{Pathological} & 0 & 9 & 272 & 2166 \\
        \bottomrule
    \end{tabular}
\end{table}

\begin{table}[H]
    \centering
    \caption{Confusion Matrix for TCN Model Predictions on Test Set (default argmax threshold)}
    \label{tab:TCN_matrix}
    \small
    \begin{tabular}{lcccc}
        \toprule
        \textbf{True/Pred} & \textbf{Noise} & \textbf{Artifacts} & \textbf{Physio.} & \textbf{Patho.} \\

        \midrule
        \textbf{Noise} & 3542 & 43 & 5 & 0 \\
        \textbf{Artifacts} & 1 & 3833 & 31 & 3 \\
        \textbf{Physiological} & 0 & 657 & 9817 & 596 \\
        \textbf{Pathological} & 0 & 47 & 205 & 2132 \\
        \bottomrule
    \end{tabular}
\end{table}

\begin{table}
    \centering
    
    \caption{Classification Performance Metrics Across Different Models. LSTM Graphoelements model based on \cite{nejedly2019graphoelements}. CNN General and CNN Specialized models based on \cite{nejedly2019artifact}.}
    \label{tab:classification_performance}
    \small
    \begin{tabular}{l ccccc}
        \toprule
        \textbf{Class/Metric} & \textbf{CNN LSTM} & \multicolumn{2}{c}{\textbf{CNN}} & \textbf{WaveNet} & \textbf{TCN} \\
        \cmidrule(lr){3-4}
        & \makecell{\textbf{Graph.}} & \textbf{Gen.} & \textbf{Spec.} & & \\
        \midrule
        \textbf{Physiological} & & & & & \\
        F1 & 0.86 & 0.90 & 0.98 & 0.96 & 0.93 \\
        Precision & 0.85 & 0.93 & 0.97 & 0.94 & 0.98 \\
        Recall & 0.87 & 0.87 & 0.97 & 0.98 & 0.89 \\
        \midrule
        \textbf{Pathological} & & & & & \\
        F1 & 0.73 & 0.64 & 0.90 & 0.90 & 0.83 \\
        Precision & 0.66 & 0.56 & 0.90 & 0.96 & 0.78 \\
        Recall & 0.82 & 0.74 & 0.91 & 0.89 & 0.89 \\
        \midrule
        \textbf{Artifacts} & & & & & \\
        F1 & 0.80 & 0.89 & 0.97 & 0.93 & 0.91 \\
        Precision & 0.85 & 0.88 & 0.96 & 0.98 & 0.84 \\
        Recall & 0.76 & 0.91 & 0.98 & 0.89 & 0.99 \\
        \midrule
        \textbf{Noise} & & & & & \\
        F1 & -- & -- & 0.98 & $\sim$1.00 & 0.99 \\
        Precision & -- & -- & 0.99 & $\sim$1.00 & $\sim$1.00 \\
        Recall & -- & -- & 0.96 & $\sim$1.00 & $\sim$0.99 \\
        \midrule
        \textbf{Macro avg.} & & & & & \\
        F1 & 0.80 & 0.81 & 0.96 & 0.94 & 0.92 \\
        Precision & 0.78 & 0.80 & 0.95 & 0.95 & 0.90 \\
        Recall & 0.82 & 0.86 & 0.96 & 0.93 & 0.94 \\
        \bottomrule
    \end{tabular}
\end{table}
\vspace{70pt}
The confusion matrix (Table~\ref{tab:WAVE_matrix}) illustrates WaveNet’s ability to discriminate noise and artifacts: noise was twice misclassified out of 3563 samples, and artifacts were seldom confused with pathological activity, but had some small recall issues with wrongly classifying physiological spikes. This is similar to the TCN baseline (Table~\ref{tab:TCN_matrix}). It indicates that WaveNet and TCN reliably identify both transient and stationary non-neural signal components. However, some misclassification persisted between physiological and pathological EEG signals, which is consistent with the known clinical challenge of differentiating subtle abnormal brain activity from normal patterns.
\clearpage

\newpage

Overall, these results highlight the WaveNet model’s balance of sensitivity and specificity across classes, achieving precision and recall values exceeding 0.93 for most categories (Table~\ref{tab:classification_performance}). This positions the model as a powerful tool for clinical iEEG analysis, with clear advantages in suppressing irrelevant variance while preserving meaningful neurophysiological signals.

\section{Discussion}
Notably, the WaveNet model was trained without patient specific tuning, yet achieved strong performance across clinically diverse iEEG datasets. This outcome shows its capacity to generalize under real world conditions marked by variability and imbalance. Such inherent flexibility points toward a broader application in synthesize long range temporal dependencies and multi channel interactions in high density neural recordings. In particular, it opens the door to applying similar architectures to Neuropixels data within spike sorting pipelines like Kilosort, where waveform diversity and spatio-temporal structure require adaptable and scalable models. Building on these principles, future work will extend this approach to neuronal spike classification in large-scale iEEG recordings.

Spike-sorting approaches are shifting from fixed template clustering toward dynamic, interpretable, and biologically informed deep learning methods. Grounded in both neuroscience and machine learning theory, this framework aims to be scalable, transparent, and adaptable. Given WaveNet's original design for raw audio synthesis, the model's architecture could also be leveraged in future iterations to generate new healthy and epileptical spike, noise, and artifact signals for brain modeling input. This framework's demonstrated generalizability positions it as a strong base model for hybrid systems, integrated with tools like Kilosort and extended via post hoc classifiers and explainability techniques. Future iterations will also incorporate tools such as Grad-CAM to elucidate model decision-making, enabling biologically meaningful interpretations of spike classifications.

Additionally, while the WaveNet model demonstrated high performance on diverse human iEEG datasets, other architectures, such as the recently proposed Attention Recurrent Neural Network (ARNN) by Rukhsar and Tiwari \cite{rukhsar2024arnn}, have reported impressive accuracy on long-duration, multi-channel seizure datasets, including CHB-MIT and canine iEEG from the Mayo Clinic and North Carolina State University. While ARNN achieves strong accuracy, particularly in per-subject evaluations, it utilized 30 epochs for training, whereas our WaveNet model’s training had a hard stop at 10 epochs due to limited computing access. Future research should directly compare WaveNet and ARNN models on shared, heterogeneous iEEG datasets, under similar training constraints, to evaluate their respective trade-offs in generalization, computational efficiency, and interpretability.

Both models leverage temporal pattern recognition for their high computational efficiency and accuracy. The ARNN model uses an LSTM-style recurrent gate to process sequences, while our WaveNet model employs dilated causal convolutions for temporal pattern recognition, followed by two convolutional ending layers. Given that traditional LSTM models, as baselines in other studies, have shown comparatively lower performance than more advanced temporal models on certain iEEG datasets, an intriguing future direction is to explore hybrid architectures. Specifically, combining WaveNet’s unique capacity for capturing long-range temporal dependencies via dilated convolutions with the sequence modeling capabilities of recurrent networks could potentially enhance the efficiency and accuracy of models like ARNN.

\section{Limitations}
Despite the high classification performance of the WaveNet model, several limitations must be acknowledged. First, the model's performance on the distinction between physiological and pathological iEEG signals, while high, is not absolute. The observed cross-class false positives (175 for physiological and 272 for pathological) suggest that the model occasionally struggles with the inherent morphological similarity between some healthy sharp transients and epileptiform activity. This reflects a broader challenge in clinical neurophysiology where definitive ground truth can sometimes be subjective.

Second, the current study utilized a fixed 3-second windowing approach. While effective for capturing the transient nature of spikes and noise, it may not fully capture the slower, evolutional temporal dynamics of longer ictal events or state-dependent changes in the background EEG. Third, due to computational resource constraints, the training was limited to 10 epochs. While the convergence curves (Fig. \ref{fig:ACC_plot} and \ref{fig:LOSS_plot}) suggest stability, a more exhaustive hyperparameter search and extended training could potentially yield further marginal gains in precision.

Finally, while the dataset includes multi-site data from the Mayo Clinic and St. Anne’s University Hospital, the model’s performance on completely "out-of-distribution" data ,such as recordings from different amplifier hardware or significantly different sampling rates, remains untested. Future work should focus on cross-dataset validation to ensure the WaveNet’s dilated convolutions have learned generalizable biomarkers rather than site-specific signatures.

\section{Conclusion}
This study demonstrates that the WaveNet architectures, originally developed for audio synthesis, can be effectively repurposed for classification of physiological, pathological, artifact, and noise in iEEG . Its capacity to capture both fine-grained waveform subtleties and long-range temporal dependencies makes it well-suited for neurophysiological data. Leveraging a merged dataset from St. Anne’s University Hospital (Brno, Czech Republic) and the Mayo Clinic (Rochester, Minnesota, USA), the model achieved high validation and test accuracies, despite significant class imbalance and inter-subject variability.

WaveNet outperformed generalized CNN models across most metrics, while the specialized CNN achieved higher performance on physiological iEEG signals (F1 0.96 vs. 0.98; Table~\ref{tab:classification_performance}). The TCN baseline trained on the same preprocessing and dataset splits achieved performance comparable to WaveNet on several classes, demonstrating that a non-autoregressive convolutional architecture can capture iEEG temporal patterns effectively. The specialized CNN reported by Nejedly et al. \cite{nejedly2019artifact} highlights the trade off of highly tuned models: they maximize performance on particular signal types but require patient-specific tuning. WaveNet provides strong, consistent performance across classes, including improved pathological classification relative to the generalized CNN (F1 0.81 vs. 0.64). Which might reflect its deeper temporal receptive field and residual connections that capture long-range or irregular dynamics in iEEG signals.

In conclusion, this study demonstrates that WaveNet can serve as a powerful and generalizable framework for iEEG analysis, capable of capturing fine grained waveform features while maintaining accuracy across heterogeneous datasets. Its capacity to synthesize long-range temporal dependencies and scale across diverse recording contexts suggest its utility for clinical applications such as seizure detection and artifact rejection. Equally important, the architecture allows integration with interpretability methods, ensuring that future implementations can prioritize transparency and biological plausibility alongside performance. Together, these contributions put WaveNet forward as an option for a flexible foundation for advancing neural signal synthesis and classification, with potential extensions to large-scale iEEG data and hybrid deep learning approaches that further interweave neuroscience and machine learning.

\section*{Acknowledgments}
I would like to thank the team at St. Anne’s University Hospital (Brno, Czech Republic) as well as the Mayo Clinic (Rochester, Minnesota, USA) for providing access to the EEG recordings that made this research possible. Special thanks go to Dr. Khubaib Ahmed for his valuable feedback on this article.
This research was conducted independently and without institutional funding. The author acknowledges the importance of interdisciplinary collaboration in bridging neuroscience and artificial intelligence, and hopes this work contributes to the development of more adaptive and interpretable neurotechnological tools.

\bibliographystyle{IEEEtran}
\bibliography{references}

% --- APPENDICES ---
\appendix

\section*{Appendix}
\section*{Detailed wavenet Training Configuration}
\label{app:training_config}
% NOTE: You refer to Figure A1 in the main text, but there is no figure here.
% You should add the figure environment for it below.
% For example:
% \begin{figure}[!t]
%   \centering
%   \includegraphics[width=0.8\columnwidth]{your_figure_a1.png}
%   \caption{Caption for your appendix figure.}
%   \label{fig:appendix_dropout}
% \end{figure}
\renewcommand{\thefigure}{\Alph{section}\arabic{figure}}

\begin{itemize}
    \item \textbf{Data Preparation:} Batches were well-mixed, with our checks confirming that 0 \% were highly clustered (defined as bigger than 90 \% class dominance within a batch). This ensured effective global shuffling and diverse batches, as demonstrated by the representative analysis of the first five batches where the maximum class dominance ranged from 40.62 \% to 68.75 \%.
    \item \textbf{Loss Function:} Custom Focal Loss with parameters \texttt{gamma=2.0} and class-weight-derived \texttt{alpha} values to address class imbalance.
    \item \textbf{Hardware and Platform:} Training was conducted on Kaggle using a Tesla P100-PCIE-16GB GPU (CUDA platform) with \texttt{cuDNN version 90300} and XLA compilation for performance.
    \item \textbf{Model Architecture:} WaveNet with 7 dilated causal convolutional layers, each with 32 filters, kernel size 3, Swish activation, and L2 regularization (\( \lambda = 0.0001 \)). Dilation rates: [1, 2, 4, 8, 16, 32, 64].
    \item \textbf{Training Duration:} Total runtime was approximately 11 hours, 19 minutes, and 16 seconds.
    \item \textbf{Hyperparameters:}
    \begin{itemize}
        \item Input sequence length: 15000
        \item Batch size: 32 (effective 64 via gradient accumulation)
        \item Learning rate: 0.001
        \item L2 regularization: 0.0001
    \end{itemize}
    \item \textbf{Adaptive Dropout:} The dropout rate was adjusted dynamically based on a composite score.
    \begin{center}
\colorbox{white}{
  \parbox{0.8\columnwidth}{
    \centering
    \small
    \textbf{Composite Score} = (PrevAcc $-$ ValAcc) $\times$ 1.5 \\[2pt]
    \phantom{\textbf{Composite Score} = }+ (PrevAUC $-$ AUC) $\times$ 0.8 \\[2pt]
    \phantom{\textbf{Composite Score} = }+ (ValLoss $-$ PrevLoss) \\[2pt]
    \phantom{\textbf{Composite Score} = }+ (RelGap $-$ BaselineGap) \\[2pt]
    \phantom{\textbf{Composite Score} = }+ 25 $/$ ValNoLearn
  }
}
\end{center}

Where:
\begin{itemize}
  \item \textbf{PrevAcc}: Validation accuracy from the previous epoch.
  \item \textbf{ValAcc}: Current epoch's validation accuracy.
  \item \textbf{PrevAUC}: Validation AUC from the previous epoch.
  \item \textbf{AUC}: Current epoch's AUC.
  \item \textbf{ValLoss / PrevLoss}: Validation loss from the current/previous epoch.
  \item \textbf{RelGap}: \( \frac{\max(\text{TrainAcc}, 10^{-8}) - \text{ValAcc}}{\max(\text{TrainAcc}, 10^{-8})} \) 
  \item \textbf{BaselineGap}: Constant (set to 0.02) defining an acceptable generalization gap.
  \item \textbf{ValNoLearn}: \( |\text{ValAcc} - \text{PrevAcc}| \) if less than 0.01, else 20.
\end{itemize}

Positive Composite Scores indicate stagnation, performance degradation, or overfitting, which increase the dropout rate via(Fig. A1):

New dropout = Current dropout + 0.05xComposite score

This adaptive mechanism encourages regularization when learning stagnates or generalization degrades. Similar to normal dropout, however automated changing after every epoch.

\end{itemize}

\begin{figure}[!htb]
    \centering
    \includegraphics[width=0.8\columnwidth]{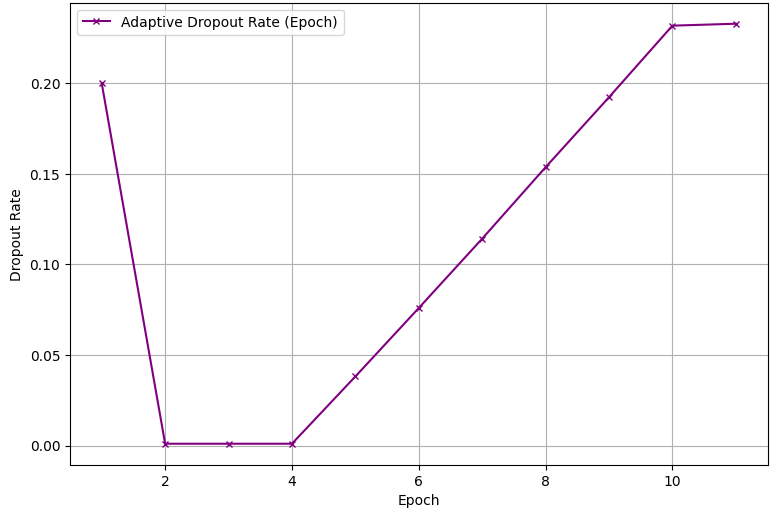}
    \caption{Dropout rate dynamics throughout training, governed by the composite performance score.}
    \label{fig:adaptive_dropout_plot}
\end{figure}

\end{document}